\def\year{2021}\relax
\documentclass[letterpaper]{article} 
\usepackage{aaai21} 
\usepackage{times} 
\usepackage{helvet} 
\usepackage{courier} 
\usepackage[hyphens]{url} 
\usepackage{graphicx} 
\usepackage{graphicx}  
\usepackage{natbib} 
\usepackage{caption} 
\usepackage{times}
\usepackage{epsfig}
\usepackage{graphicx}
\usepackage{amsmath}
\usepackage{amssymb}
\usepackage{calligra}
\usepackage{calrsfs}
\usepackage{mathtools}
\usepackage{url}
\usepackage{booktabs}
\usepackage{multirow}
\usepackage{siunitx}
\usepackage{xcolor}
\usepackage{titleref}

\title{Robust PDF Document Conversion \\ Using Recurrent Neural Networks}

\author {
  Nikolaos Livathinos,
  Cesar Berrospi,
  Maksym Lysak,
  Viktor Kuropiatnyk, \\
  Ahmed Nassar,
  Andre Carvalho,
  Michele Dolfi,
  Christoph Auer,
  Kasper Dinkla,
  Peter Staar
  \\
}

\affiliations {
    IBM Research\\
    S\"aumerstrasse 4\\
    8803 R\"uschlikon, Switzerland \\
    \{nli, ceb, mly, vku, ahn, val, dol, cau, dkl, taa\}@zurich.ibm.com \\
}

\begin{document}

\maketitle

\begin{abstract}
The number of published PDF documents in both the academic and commercial world has increased
exponentially in recent decades. There is a growing need to make their rich content discoverable
to information retrieval tools. Achieving high-quality semantic searches demands that a document's
structural components such as title, section headers, paragraphs, (nested) lists, tables and
figures (including their captions) are properly identified. Unfortunately, the PDF format is known
to not conserve such structural information because it simply represents a document as a stream of
low-level printing commands, in which one or more characters are placed in a bounding box with a
particular styling. In this paper, we present a novel approach to document structure recovery in
PDF using recurrent neural networks to process the low-level PDF data representation directly,
instead of relying on a visual re-interpretation of the rendered PDF page, as has been proposed in
previous literature. We demonstrate how a sequence of PDF printing commands can be used as input into a
neural network and how the network can learn to classify each printing command according to its
structural function in the page. This approach has three advantages: First, it can distinguish
among more fine-grained labels (typically 10--20 labels as opposed to 1--5 with visual methods),
which results in a more accurate and detailed document structure resolution. Second, it can
take into account the text flow across pages more naturally compared to visual methods because it
can concatenate the printing commands of sequential pages. Last, our proposed method needs less
memory and it is computationally less expensive than visual methods. This allows us to deploy such
models in production environments at a much lower cost. Through extensive architectural search in combination
with advanced feature engineering, we were able to implement a model that yields a weighted average
$F_1$ score of 97\% across 17 distinct structural labels. The best model we achieved is currently served in production environments
on our \textit{Corpus Conversion Service} (CCS), which was presented at KDD18. This
model enhances the capabilities of CCS significantly, as it eliminates the
need for human annotated label ground-truth for every unseen document layout. This proved particularly useful when applied
to a huge corpus of PDF articles related to COVID-19.
\end{abstract}

\begin{table*}[t]
\centering
\begin{tabular}{|l|c|cccccccccc|}
\toprule
\rotatebox[origin=l]{0}{model}
& \rotatebox[origin=l]{90}{dataset}
& \rotatebox[origin=l]{90}{title}
& \rotatebox[origin=l]{90}{author}
& \rotatebox[origin=l]{90}{affiliation}
& \rotatebox[origin=l]{90}{abstract}
& \rotatebox[origin=l]{90}{subtitle}
& \rotatebox[origin=l]{90}{text}
& \rotatebox[origin=l]{90}{picture}
& \rotatebox[origin=l]{90}{table}
& \rotatebox[origin=l]{90}{caption}
& \rotatebox[origin=l]{90}{list}
 \\ \hline
 \multirow{2}{*}{RRF} & uniform   & 0.93 & 0.98& 0.96& 0.98& 0.95& 0.99& 0.97 & 0.99 & 0.97 & -- \\
    & diverse       & 0.71 & 0.75& 0.77& 0.75& 0.86& 0.97& 0.88 & 0.97 & 0.88 & 0.65 \\
\hline
 \multirow{2}{*}{\begin{tabular}[l]{@{}l@{}}Mask-RCNN\\ \quad cut-off=0.7\end{tabular}} &
uniform   & \color{gray} N/A   &\color{gray} N/A   &\color{gray} N/A    &\color{gray} N/A   & 0.47 & 0.96& 0.59 & 0.97 &\color{gray} N/A & -- \\
& diverse   &\color{gray} N/A   &\color{gray} N/A   &\color{gray} N/A    &\color{gray} N/A   & 0.57 & 0.92 & 0.83 & 0.92 &\color{gray} N/A & 0.13 \\

\hline
 \multirow{2}{*}{\begin{tabular}[l]{@{}l@{}}Mask-RCNN\\ \quad cut-off=0.95\end{tabular}} &
uniform   &\color{gray} N/A   &\color{gray} N/A   &\color{gray} N/A    &\color{gray} N/A   & 0.61 & 0.97& 0.76 & 0.98 &\color{gray} N/A & -- \\
& diverse   &\color{gray} N/A   &\color{gray} N/A   &\color{gray} N/A    &\color{gray} N/A   & 0.72 & 0.94& 0.87 & 0.8 &\color{gray} N/A & 0.2 \\

\hline
\begin{tabular}[l]{@{}l@{}}Seq2Seq model\\ \quad (this paper)\end{tabular}    & diverse       & 0.96 & 0.95& 0.90& 0.95& 0.87& 0.98& 0.95 & 0.94 & 0.96 & 0.70 \\
\bottomrule
\end{tabular}
\centering
\caption{\label{table:sota} Evaluation results of state-of-the-art AI-based PDF structure recovery methods.}
\end{table*}

\section{\label{sec:Intro}Introduction}

The Portable Document Format (PDF) was introduced in 1993 and has been widely adopted throughout the
academic and commercial world. In 2015, Adobe estimated ~\cite{billionsPdfs} that there were 1.5
trillion PDF documents in circulation. That number has certainly risen since then.

At its core, the PDF format encapsulates a printing language. Practically, this means that a PDF
document is a set of consecutive printing instructions to place one or more characters, lines and
bitmap images in a certain position on the page with a specific style. As a consequence, the
structural context of the characters/lines within the document is lost.  Without visual
interpretation of the page, it is very difficult to deduce whether a specific set of characters from
a printing command belongs to a title, abstract, table, etc. This is particularly problematic when
one wants to retrieve and analyze the content of the document. Without knowledge of the document
structure, one can only perform a basic keyword search, which is very limiting. To extract content
from documents accurately, one needs to know the structure of the document, which in turn requires
robust and accurate conversion of PDF documents into a well-defined data layout such as JSON or
HTML. The latter remains very challenging still today.

These limitations of the PDF format have inspired many researchers to build image segmentation
algorithms that can detect document components in a visual rendering of a page. In particular,
object detection methods based on deep neural networks have been extensively used to find particular
document components such as figures or tables \cite{DeepFigures, Publaynet}. The disadvantage of
using image-based object detection methods is that they often fail to predict sufficiently accurate
regions required for high-quality structure recovery. Furthermore, they are computationally
expensive. This makes them impractical for the task of robust document conversion at scale.

In this paper, we present a new method to support robust PDF document conversion. The novelty of
this method is that we do not rely on image segmentation methods. Rather, we apply neural network
architectures originally designed for natural language processing (NLP) tasks to the sequence of
printing commands in the PDF code. This work was inspired by recent advances in NLP. We investigated
whether we can replace the input embeddings of characters or words in traditional NLP networks with
the features of the PDF printing commands. This yields a classification for each PDF printing
command according to a fixed set of labels describing a document layout feature. Examples of such
labels are \textit{title}, \textit{author}, \textit{affiliation}, \textit{abstract},
\textit{subtitle} (of different levels), \textit{text}, \textit{list} (including nested lists),
\textit{table}, \textit{figure}, \textit{reference}, \textit{caption},~etc.

\section{\label{sec:SOTA}Previous Related Work and State of the Art}
There exist several approaches to the problem of document structure recovery with different types of
outputs, some of which are implemented in commercial product offerings for large-scale document
conversion (e.g.~Amazon Textract, Docparser, IBM Watson Discovery). Broadly speaking, one can
distinguish between two categories of approaches, which we will refer to as \textit{image-based
structure recovery} and \textit{PDF representation-based structure recovery}.

In image-based structure recovery, algorithms or models are typically designed to detect specific
document components in a visual rendering of each page. Representatives of this type are the work
presented in the DeepFigures paper \cite{Siegel2018ExtractingSF} and the PubLayNet paper
\cite{Publaynet}, which target the extraction of tables, figures, headers (comprising entities such
as title, chapter or subtitle), text blocks and lists.
Whereas delivering generally favorable results for documents with an unseen layout, such approaches
have fundamental problems: Firstly, in many instances, the segmentation quality is not sufficient
for high-quality document conversion. It has been reported that the $F_1$ score for detecting tables
and images with the DeepFigures model (based on ResNet101) is in the range of 80--85\% for an
expected intersection over union (IOU) of $\ge$80\%. For PublayNet, a mean average precision (mAP)
of 90--91\% was reported for an IOU of $\ge$50\%. There is still a significant risk that elements in
a document page are misclassified or entirely missed because they are not included in any segmented
region or detection box.

Secondly, the detail resolution of the detected document components is limited due to the problem
that training on fine-grained segmentation labels (e.g.~distinguishing among different list nesting
levels, abstract, caption, subtitles) for visually ambiguous elements is prone to create label
confusion and overlapping detections.

In PDF representation-based structure recovery, each individual text cell (generated by one PDF
printing command) is classified according to a fixed set of labels by considering only those
features that are derived from their data representation and ordering in the PDF code. This allows a
more detailed structure recovery. The Recursive Random Forest (RRF) models used in Corpus Conversion
Service (CCS), \cite{KDD2018} fall in this category and predict fine-grained labels on the order of
10--20 such as subtitles of different depths, nested lists, page headers or footers, footnotes,
bibliographic data, etc. As these models predict a label for each text cell, they allow us to
evaluate their performance based on metrics derived from a label confusion matrix, as we showed in
detail in our previous work \cite{KDD2018}. We are unaware of any existing, deep-learning-based
approach that works directly on the native representation of PDF text cells.

In order to validate our newly proposed document-structure recovery approach with the state of the
art, we establish two baselines:
\begin{itemize}
  \item Performance of a trained CCS RRF model, computed by a fivefold cross-validation on a uniform
      layout dataset (\textit{uniform}) and a \textit{diverse} layout dataset (see
        \titleref{sec:dataset}). The RRF model is trained on ten labels, namely
        title, abstract, author, affiliation, subtitle, text, picture, table, caption and list. The
        $F_1$ score is computed directly from the confusion matrix over the labels assigned to each
        text cell.
  \item Performance of a Mask-RCNN network trained on PubLayNet dataset with Facebook's Detectron
      framework\footnote{https://github.com/facebookresearch/detectron2}, then applied to the diverse
	  layout dataset (full) for region segmentation. In order to compute the $F_1$ score over
	  labels of text cells, we run bitmap images of all document pages (1025 by 1025 pixels)
	  through the pre-trained Mask-RCNN network, retrieve the predicted segmentation masks for
	  five labels defined in PubLayNet, namely subtitle, text, picture, table and list, and assign
	  to each text cell the label of the segmentation mask in which they are included to at least
	  80\%,  otherwise they are not assigned a label. The confidence level at which a region
	  prediction is accepted is controlled by a cutoff value of either 70 or 95\%.
\end{itemize}

Table~\ref{table:sota} shows the $F_1$ scores achieved by the RRF and Mask-RCNN models. The RRF
model performs extremely well on a dataset with uniform layout (\textit{uniform}), as was reported
in \cite{KDD2018}. On the \textit{diverse} dataset containing multiple layouts, the classification
performance of the RRF degrades significantly. The RRF models are simply not robust enough for us to
learn the structure for multiple layouts at once. The results of the Mask-RCNN model are mixed: It
generally has good performance for text-block and table detection, but only average performance for
subtitles and pictures. Lists are recalled poorly. For inference, Mask-RCNN occupies close to 2~GB
of GPU memory and takes 120~ms per page image on an NVIDIA Tesla P100~GPU.  For comparison, we also
show the $F_1$ scores achieved by our best seq2seq model developed and presented in this paper
(bottom). Its strength is clearly that it performs very well for all labels over the entire
\textit{diverse} dataset.

\section{\label{sec:Networks}Networks, Datasets and Evaluation}

This section presents our new approach to PDF document conversion. Additionally, we document how we
evaluate the newly developed models and describe in detail the datasets we constructed and used in
the evaluation.

\subsection{Networks}

\begin{figure}[b!]
\centering
\includegraphics{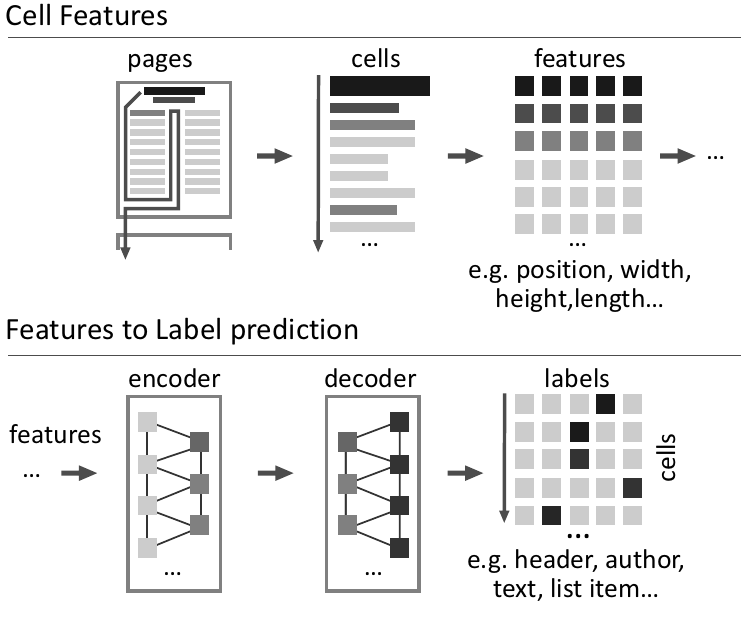}
\caption{\label{fig:sketch_network} Sketch of the network architecture for a generic page model. In
    each model, we use the sequence of features of each text cell as input. The ordering of the
    cells is obtained after sorting them according to reading order using the toposort algorithm.
    The output of the network yields a label classification for each cell. Each model with which we
    have experimented contains at least the encoder part. The embedding and decoder parts are
    optional components.}
\end{figure}

The natural flow of the text in a document is very often reflected in a sequence of PDF printing
commands. This inspired us to look at recurrent neural networks and, in particular, networks used
commonly in NLP. Their \textit{unreasonable effectiveness} (see \cite{Sejnowski201907373}) in
domains such as \textit{named entity recognition} \cite{lample-etal-2016-neural}, \textit{machine
translation} \cite{klein-etal-2017-opennmt} and \textit{chemistry} \cite{C8SC02339E} clearly
demonstrates that they can capture the underlying structure and trends in (noisy) sequences of data.
In the case of traditional NLP, the signals are often the embeddings of each character or word in a
string. In chemistry applications, the signals are the embeddings of the characters in the SMILES
representation of a chemical compound.

For PDF document conversion, we use a set of features associated with each text cell as the input
signal. In this context, a text cell is the final output of a PDF printing command. Each text cell
provides its lower-left coordinate $\{x_0, y_0\}$, its width \& height, its textual content and its
text style. As illustrated in Fig.~\ref{fig:sketch_network}, an ordered sequence of text cells
produces a multi-dimensional signal, which comprises the $\{x_0, y_0\}$ coordinate, width and height
of the text cell and potential derived quantities (e.g.~horizontal/vertical distance to
previous/next cell). The dimension of this signal is equal to the number of features associated with
each text cell. The length of the signal is obviously equal to the number of text cells of the
document. The goal of the network is to assign a label to each cell. In the most generic case, this
can be accomplished by feeding the previously described multi-dimensional signal as an input to a
signal-processing network. The latter will then encode these signals. In traditional
sequence-to-sequence methods, one then uses a decoder, followed by a linear transformation and a
softmax function to obtain a label prediction for each cell in the sequence. The use of a decoder
here is optional because the length of the input signal is always the same as the length of the
output signal. As such, one can in principle skip the decoder step and directly apply the linear
transformation followed by the softmax function to obtain a label prediction for each cell in the
sequence. See \titleref{sec:results} for a detailed exploration of these different options.

\subsection{\label{sec:dataset}Datasets}

The training data used for our models was obtained through human annotation of PDF documents on the
CCS platform  as documented in detail in \cite{KDD2018}. In essence, we annotated 2,940 PDF pages
covering seven different publishing layouts (e.g.~Elsevier, Physical Review) using 17 different
labels. We refer to this as the \textit{diverse} dataset. A single-layout subset of this dataset,
which only contains articles from the \textit{Physical Review} journal, is referred to as a
\textit{uniform} dataset (see \titleref{sec:SOTA}).
For reproducibility, we have transformed all parsed PDF pages in the dataset, including their
annotations into easy-to-consume JSON files and published them on
Kaggle\footnote{https://www.kaggle.com/peterstaar/pdf-document-conversion-dataset}. In each JSON
file, we provide the features (geometric and textual) of each text cell (or PDF printing command) as
well as its associated label. All geometric features are renormalised relative to the page
dimensions.

\begin{figure}[t]
\centering
\includegraphics{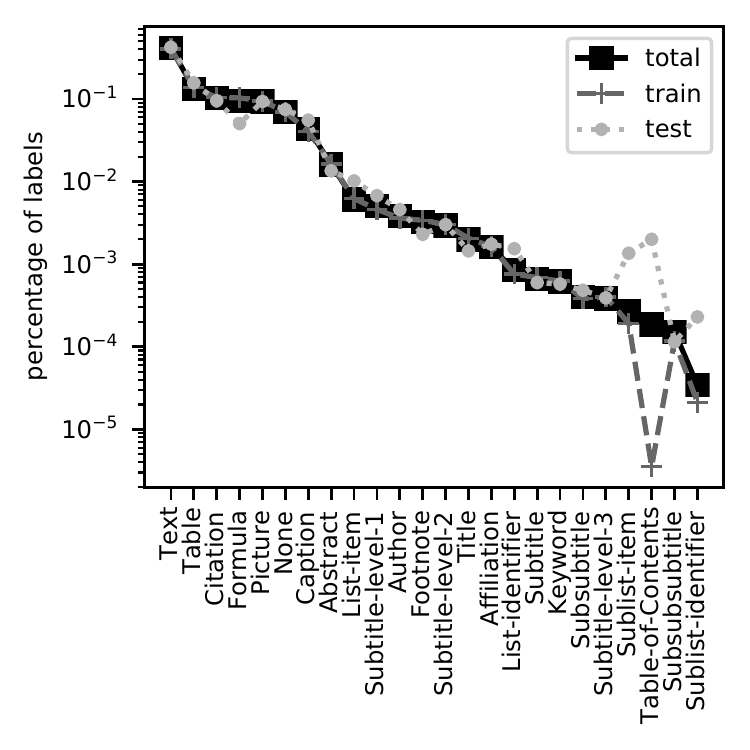}
\caption{\label{fig:label_dist}  Label distribution across the entire data set (see
	\titleref{sec:dataset}). The plot shows the distribution of the labeled cells in the entire
    data set as well as in the training and test sets.}
\end{figure}

As shown in Fig.~\ref{fig:label_dist}, the distribution of the labels in the dataset is heavily
skewed. This is expected because the number of cells with the label \textit{text} is obviously much
greater than that with the label \textit{title}. Therefore, special care must be taken to split the
data into training and test sets for representative and reproducible evaluation scores. This can be
achieved by ensuring that the distribution of the labels is similar in both sets. Traditional 90--10
random splitting of the dataset into training and test sets is not recommended because this often
does not result in a similar distribution of labels.

To obtain training and test sets with similar label distributions, we took two actions. First, we
made the test set relatively large by choosing a 60--40 split of the data into training and test
sets. Second, we distributed the pages across the training and test sets in a controlled way. More
precisely, we created a mapping from each label to the pages in which the labels occur. This allowed
us to rank the labels according to the number of pages in which they occur, as shown in the
left-hand part of Fig.~\ref{fig:label_dist}. Next, we iterated over the labels, starting with the
label with the least number of pages and progressively visited the labels that occur more often. For
each label iteration, we first removed all the pages that had already been assigned in the training
and test sets. The remaining pages for that label were then split according to the train-test ratio
(e.g.~60--40) and assigned in their set accordingly. This procedure allowed us to progressively
split the pages into a training and a test set while ensuring that the distribution of labels is
similar for both sets, as shown in Fig.~\ref{fig:label_dist}. For simplicity, we have added a few explicit
training and test-set mappings to the Kaggle dataset, both for the entire dataset as well as for each of
the seven layouts.

\subsection{Evaluation}

As we are training a supervised classifier, all evaluation metrics can be derived from the confusion
matrix over the cell labels. In order to compare and ultimately rank models with different network
architectures in the next section, we need to obtain a single number from the confusion matrix. We
chose to compute the $F_1$ average weighted by support, in which the weights are proportional to the
occurrence frequency of each label, to account for the label imbalance.

\begin{table*}[t]\centering
\begin{tabular}{|l|cc|ccc|ccc|ccc|}
\toprule
& \rotatebox[origin=l]{90}{\begin{tabular}[l]{@{}l@{}}feeding\\ mode\end{tabular}}
& \rotatebox[origin=l]{90}{\begin{tabular}[l]{@{}l@{}}active\\ decoder\end{tabular}}
& \rotatebox[origin=l]{90}{$k$}
& \rotatebox[origin=l]{90}{$d$}
& \rotatebox[origin=l]{90}{$h$}
& \rotatebox[origin=l]{90}{avg. $F_1$}
& \rotatebox[origin=l]{90}{avg. $R$}
& \rotatebox[origin=l]{90}{avg. $P$}
& \rotatebox[origin=l]{90}{time [msec]}
& \rotatebox[origin=l]{90}{size [mbyte]}
& \rotatebox[origin=l]{90}{size [param]}
 \\ \hline \hline
\textit{ model-0 } &  direct  &  yes  & 1 &  128  & -- & 0.05  &  0.05  &   0.05  & 151  & 0.7 & 105,817 \\
\textit{ model-0 } &  indexed  &  yes  &  2  &  64  & -- & 0.25  &  0.42  &  0.18  &  148  & 1.8 & 385,241 \\ \hline
\textit{ model-1 } &  indexed  &  yes  & 2 &  64  & -- & 0.35  &  0.42  &  0.39  &  407  &  5.2 & 426,713 \\
\textit{ model-1 } &  indexed  &  no  & 2 &  64  & -- & 0.79  &  0.79  &  0.79  &  216  &  4.2 & 340,569 \\ \hline
\textit{ model-2 } &  indexed  &  yes  & 2 &  64  & -- & 0.54  &  0.55  &  0.60  &  575  &  8.6 & 725,273 \\
\textit{ model-2 } &  stacked  &  yes  & 2 &  64  & -- & 0.59  &  0.60  &  0.60  &  355  &  7.7 & 642,713 \\
\textit{ model-2 } &  indexed  &  no  & 2 &  64  & -- & 0.83  &  0.84  &  0.83  &  397  &  5.6 & 457,881 \\
\textit{ model-2 } &  stacked  &  no  & 2 &  64  & -- & 0.96  &  0.96  &  0.96  & 249 &  5.4 & 443,545 \\ \hline
\textit{ model-3 } &  indexed  &  yes  & 2 &  64 & -- & 0.51  &  0.53  &  0.51  &  2,324  &  5.5 & 456,217 \\
\textit{ model-3 } &  indexed  &  no  & 2 &  64 & -- & 0.81 &  0.82  &  0.81  &  612  &  4.4 & 350,489 \\
\textit{ model-3 } &  stacked  &  no  & 2 &  64 & -- & 0.96 &  0.96  &  0.96  &  501 &  2.8 & 215,321 \\ \hline
\textit{ model-4 } &  indexed  &  yes  & 2 &  64 & -- & 0.57  &  0.58  &  0.60  &  3,729  &  9.8 & 830,233 \\
\textit{ model-4 } &  indexed  &  no  & 2 &  64 & -- & 0.85  &  0.85  &  085  &  714  &  6.0 & 494,105 \\
\textit{ \textbf{model-4}} & \textbf{stacked} &  \textbf{no} & \textbf{2} &  \textbf{64} & \textbf{-} & \textbf{0.97} & \textbf{0.97}  & \textbf{0.97} & \textbf{496} & \textbf{5.8} & \textbf{479,769} \\ \hline
\textit{ model-5 } &  indexed  &  no  & 2 &  200 & 4 &  0.59  &  0.62  &  0.60  &  63  &  10.7 & 889,025 \\
\bottomrule
\end{tabular}
\caption{\label{table:models} Accuracy and performance results for different models. The networks
    considered here are \textit{model-0} (seq2seq with uni-directional GRU layers and no attention),
    \textit{model-1} (seq2seq with uni-directional LSTM layers and no attention),  \textit{model-2}
    (seq2seq with bi-directional LSTM layers and no attention), \textit{model-3} (seq2seq with
    uni-directional LSTM layers and attention), \textit{model-4} (seq2seq with bi-directional LSTM
    layers and attention) and \textit{model-5} (transformer model). Columns $k$, $d$ and $h$
    represent the optimal number of layers, the encoding dimension and, for \textit{model-5}, the
    number of heads, respectively. The optimal hyper-parameters were obtained from a hyper-parameter
    search with $k \in {1, 2, 4}$, $d \in {32, 64, 128, 256}$, and $h \in {2, 4, 8}$. All models
    have been trained and evaluated with an NVIDIA Tesla P100 GPU.}
\end{table*}

\section{\label{sec:results}Results}

This section describes our journey to find a neural network architecture with good classification
performance for text-cell labeling. The evaluation of the networks considers their accuracy
according to the weighted $F_1$ score over the labels. We also review the performance with regard to
inference time and the complexity with the overall memory footprint and the number of trainable
parameters. Not all architectures we considered were very successful. However, we learned valuable
lessons from each new network architecture, and we believe that these insights might be very useful
to the reader in the context of similar types of problems.

The search for a good model was dominated by two questions. First, which network architecture
provides good accuracy and performance? Second, what feature engineering can we apply in order to
improve the accuracy for the best-performing network architectures? We address both questions in the
following sections.

\subsection{\label{network_search}Network Architecture Search}

The impressive results of a simple sequence-to-sequence (seq2seq) architecture with a single GRU
layer\footnote{https://pytorch.org/tutorials/intermediate /seq2seq\_translation\_tutorial.html} for
machine translation motivated us to start our model search with this particular architecture. For
simplicity, we will refer to this architecture as \textit{model-0$_k^d$}, which consists of $k$ GRU
encoding layers with encoding dimension $d$ and a single GRU decoding layer. The input for this
model can be generated in two distinct modes: using the features directly (\textit{direct} mode) or
using an embedding layer (\textit{indexed} mode) as depicted in~\ref{fig:sketch_network}. In the
indexed mode, we discretize each feature into $N$ bins and construct an index lookup. The index $i$
is computed directly from the bin number $b_l$ of each feature $l$, i.e.~$i=\sum_{l=1}^{M}
N^{l-1}\,b_l$, where $M$ represents the total number of features. In this way, a sequence of cell
features can be translated into a sequence of integer numbers. The latter can then be fed into the
network with an embedding matrix as the first layer. This indexed method resembles more closely the
classical NLP networks, which start with an embedding layer for the individual characters or words
in the vocabulary. The indexed mode has a disadvantage, namely that the size of the lookup
dictionary grows exponentially with the number of features. Therefore, we restrict the input
features of the network to the geometric features $x_0$, $y_0$, width and height with a binning
equal to $N=20$. In this section, we will consider only these features. In the next section, we will
propose methods to circumvent the exponential increase of the index range in order to add more
features.

We evaluated \textit{model-0$_k^d$} for different depths (1, 2 and 4), different encoding dimensions
(32, 64, 128 and 256) and different feeding modes (direct or indexed). The best overall result was
found for indexed feeding mode, two layers and 64 encoding dimensions, resulting in an average $F_1$
of 0.25. As shown in Table~\ref{table:models}, the results for \textit{model-0} in indexed mode were
much better than in direct mode. This was valid for any combination of the evaluated
hyper-parameters $k$ and $d$. We believe that the direct feeding mode does not perform well because
the absolute variation of feature elements might be too small. The indexed method overcomes this
issue by binning the features. Owing to the large discrepancy with regard to accuracy between the
direct and indexed feeding modes, we decided to stick to the indexed feeding mode for our further
network search.

\begin{figure*}[t]
\centering
\includegraphics{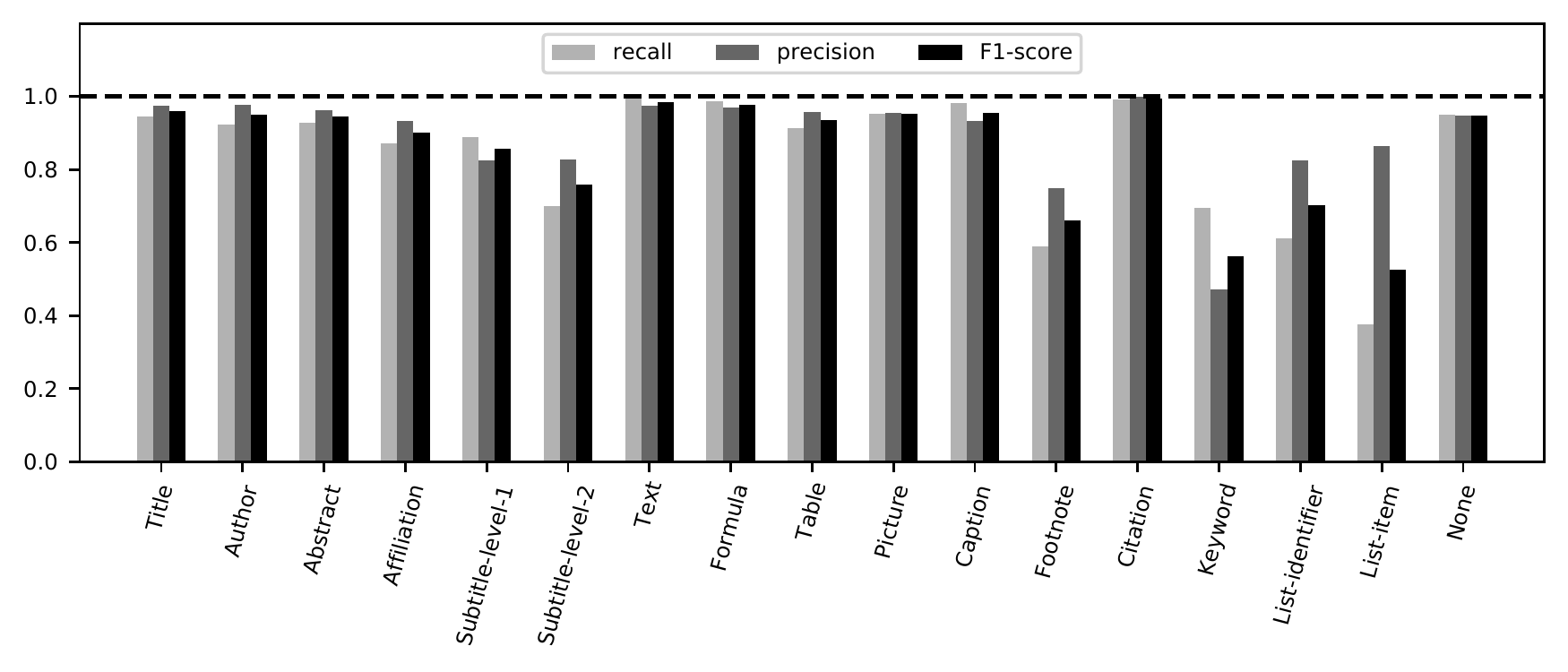}
\caption{\label{fig:best_pred} Recall, precision and $F_1$ score for all 17 labels in our dataset,
    obtained from model-4 with a stacked feeding mode, no active decoder, two bidirectional LSTM
    layers and attention.}
\end{figure*}

After inspecting the results with \textit{model-0}$^{d}_{k}$, we identified three strategies to
improve the accuracy. The first strategy was to replace the simple GRU layers with LSTM layers. LSTM
layers are known to capture complex structure much better and are used extensively in other
machine-learning tasks. The second strategy consisted of using bidirectional encoding layers, such
that the information of the cells can propagate also in reverse reading order, allowing the model to
exploit the maximum amount of information. The third strategy incorporated an attention layer, which
introduces all-to-all correlations between the cells on the page. These three strategies can be
investigated independently (i.e.~GRU versus LSTM; uni-directional versus bi-directional; attention
versus no-attention) for different numbers of layers and different encoding dimensions. However, in
order to keep the computational cost acceptable, we chose to drop the GRU option. Therefore, we
created four new models, namely \textit{model-1}$^{d}_{k}$ (seq2seq with uni-directional LSTM layers
and no attention),  \textit{model-2}$^{d}_{k}$ (seq2seq with bi-directional LSTM layers and no
attention), \textit{model-3}$^{d}_{k}$ (seq2seq with uni-directional LSTM layers and attention) and
\textit{model-4}$^{d}_{k}$ (seq2seq with bi-directional LSTM layers and attention).

After evaluating each model for a different number of layers ($k \in \{1, 2, 4\}$ and encoding
dimensions $d \in \{32, 64, 128, 256\}$), we found a significant increase in the accuracy of the
models. As shown in Table~\ref{table:models}, the LSTM-based models increased the overall accuracy
by at least $\sim 10$ percentage points compared to the GRU-based model in indexed feeding mode.

During the training of model-1, -2, -3 and -4, we noticed that a significant portion of the time was
spent on the decoding layer. This is mostly due to the fact that the decoder needs its own output of
the previous iteration and is therefore inherently sequential in nature. This sequential nature of
the decoder prevents parallelism during the forward pass---which is one of the reasons why
transformers were introduced---and makes it slow. For seq2seq models with variable sequence lengths
for input and output, this approach is appropriate. However, in our case, the length of the output
sequence is equal to the length of the input sequence. Therefore, we can omit the dynamic for-loop
in the decoding layer and directly apply the softmax to the encoded cells. Surprisingly, we found
that the models without this \textit{active} decoder have a better accuracy in combination with a
much better time-to-solution performance. As  can be seen in Table~\ref{table:models}, the $F_1$
scored increased on average by 30 percentage points, and the time-to-solution is several integer
factors faster.

The success of these various seq2seq models also motivated us to look into the use of
transformer-based models. Our \textit{model-5}$^{d}_{k,h}$ is an adaptation of the original
transformer architecture \cite{NIPS2017_7181}, which eliminates the transformer decoder. More
specifically, the input features (direct or indexed plus embedding) are combined with a sinusoidal
positional encoding and fed into a multi-layer transformer encoder. The last layer of the encoder is
linked to a fully connected layer that provides the final output. To our surprise,
\textit{model-5}$^{d}_{k,h}$ delivers significantly inferior results compared to the other models,
as displayed in Table~\ref{table:models}. Moreover, we observed that the model is very sensitive to
the fine-tuning of the hyper-parameters, with $F_1$ values ranging between 0.01 and 0.59. We assume
that the transformer models contain too many variables to be trained effectively on our dataset.

The final accuracy of $0.85$ for \textit{model-4} was satisfactory, considering that we only
provided a stream of indexed raw features with no augmentation. As we know that we can encode quite
a bit more information in the features (e.g.~distance to previous/next cell, ratio of
alpha-numerical characters versus alphabetic characters), we suspected that accuracy could be
improved significantly by using better data features.

\subsection{\label{feature_engineering}Feature Engineering and Embedding Strategies}

The models we presented in the previous section were built using bare geometric features only,
either in direct or indexed mode. This is suboptimal, according to our previous experience with RRF
models. In this section, we present different strategies to extend the feature set that is fed into
the models. These extended features complement the derived geometric features (e.g.~distance to
previous and next cell) and leverage textual features of the cell (e.g.~number-of-characters,
\textit{starts-with-capital}, \textit{is-bold}).

The indexed feeding mode turned out to be very successful. This is not really surprising because the
model architectures with which we experimented originate from NLP tasks, where it is common to map
characters or words into $d$-dimensional vectors via an index lookup in the embedding matrix. As
discussed above, we achieve the same result by discretizing each feature into $N$ bins (more
specifically 20 in our case) and computing its associated linear index $i=\sum_{l=1}^{M}
N^{l-1}\,b_l$, where $b_l$ is the bin index of feature $l$. It follows directly from this formula
that the range of the look-up index grows exponentially with an increasing number of features
($=M$). As such, this direct approach of discretization is not feasible for a large number of
features (e.g.~$M \approx 20-50$). In order to prevent this exponential increase, we explored the
idea of stacking (or concatenating) the discretization vectors. Concretely, this means that, for
every value $x_i$ of the feature vector $\vec{f} = \{x_1, x_2, ..., x_M\}$, we created a one-hot
encoded vector $\vec{v}_i$ across its bins. Each of these one-hot encoded vectors $\vec{v}_i$ (of
size $N$) are then stacked to create a new feature vector of size $M*N$. Therefore, a feature vector
$\vec{f}$ of size $M$ will be transformed into a binary feature vector of size $N*M$. This stacking
approach is much more practical for two distinct reasons. First, it allows us to add many features
without exponentially increasing the index range. Second, we no longer need an explicit embedding
layer because the original feature vector is transformed in an expanded feature vector in a higher
dimensional space. Hence, the memory footprint of the model is also reduced by eliminating the
expensive embedding layers.

The features of the input signal in our approach are equal to those in the RRF models, described in
detail in~\cite{KDD2018}. These include geometric features and character-level features mentioned
above. To ensure that our conversion models are language-agnostic (and can therefore be used on
documents in different languages), we do not include any actual text content.

The inclusion of the extra features in stacked feeding mode improved the accuracy by another 10--15
percentage points. We obtained an average $F_1$ score of 0.96 for model-2 and model-3, and 0.97 for
model-4. For the latter, Figure~\ref{fig:best_pred} shows the recall, precision and $F_1$ score for
all 17 labels in our dataset. The numbers in Figure~\ref{fig:best_pred} complement the numbers
presented in Table~\ref{table:sota}. Extremely prevalent labels (e.g.~text, table, formula, picture,
caption and citation) are significantly well captured (i.e.~$F_1>0.95$). We also achieve very good
evaluation numbers (i.e.~$F_1>0.9$) for bibliographic data-related labels (e.g.~title, abstract,
affiliation and authors).

\section{Model Development, Use and Deployment \\ in Production}

The AI model presented in this paper is currently used in the production version of the Corpus
Conversion Service \cite{KDD2018}. CCS is a cloud-native application that leverages several AI
models (e.g. for OCR, table and figure detection, layout classification and segmentation) for
accurate conversion of PDF documents into a richly structured JSON. Since 2016, CCS has been in use
by IBM internally and for selected key clients. In 2019, CCS became a commercial offering in IBM's
Watson Discovery Service under the brand name Smart Document
Understanding\footnote{\url{https://cloud.ibm.com/docs/discovery?topic=discovery-sdu}}.

Prior to the development of the model presented in this paper, CCS offered two models for structure
recognition in documents. One is the Mask-RCNN-based object-detection model, which predicts clusters
of text, tables, figures and headings in a given page out-of-the-box but provides little structural
detail. The other model is the RRF model, which can classify text cells in much higher detail but
requires initial user-training on every new document collection and works well only on document
collections with mostly uniform layout (see \titleref{sec:SOTA}). The presented \textit{seq2seq} model
bridges the gap between the automatic but coarse conversion of the object-detection model and the
detailed but annotation-hungry RRF models. As long as the \textit{seq2seq} model has been
pre-trained on a diverse set of documents, it enables the CCS user to immediately convert PDF
documents with high accuracy, without the need of manual annotation for every new corpus of
documents.

\begin{figure}[t]
\centering
\includegraphics[scale=0.26]{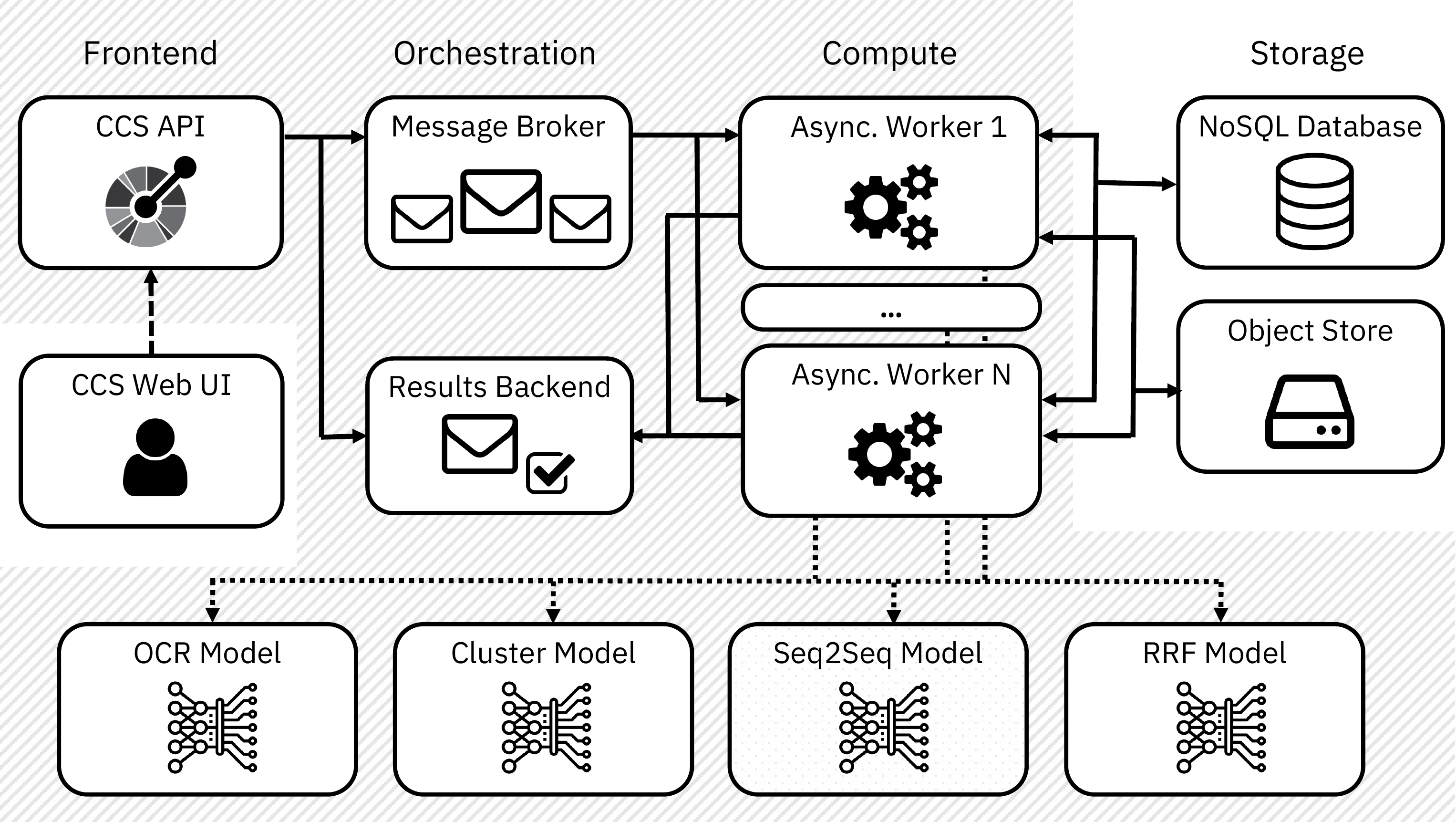}
\caption{\label{fig:sketch_deployment} Overview of the CCS platform architecture. User-actions
    triggered from the WebUI are handled by an API service, which can dispatch workloads
    asynchronously to the compute infrastructure through a queueing mechanism. Workers pick up
    queued tasks from a message broker and store the results for later retrieval on a document
    database and cloud object storage. The AI models are served as separate microservices, which
    are consumed by workers when executing the prediction pipeline. The background hatch pattern
    encloses such CCS components that run and communicate within the Kubernetes cluster environment.
    }
\end{figure}

Since the initial launch of the service in 2016, we have gathered
annotation data through the CCS platform for the training of the RRF models. After filtering out all
confidential and client related data, we obtained a training dataset of 3,695 pages, which is used in the
models listed in Table~\ref{table:models}. During the network search, we developed and trained the models
offline on a large IBM Power9 based HPC cluster.
To this point, the model architecture has converged to \textit{model-4} with stacked features and no
active decoder. Consequently, the re-training of the best model has since been semi-automated. A new
training is launched once a new ground-truth data source is vetted and accepted.

As a cloud-native application, CCS runs in a Kubernetes cluster environment
which hosts all application and service components (see Figure~\ref{fig:sketch_deployment}). All platform
capabilities are exposed through a top-level REST API service, which is both backing the CCS Web
User Interface and consumed in third-party automation pipelines and products.
Compute workloads, such as parsing PDF into an internal representation, applying the AI model
pipeline, and transforming output into JSON, are chunked into equal batches of several PDF pages and
distributed asynchronously to multiple worker pods through a message queue. As such, each workload
type can be scaled horizontally across cluster nodes to maximise throughput with the available resources.

With the introduction of the \textit{seq2seq} model presented here, all AI models (in particular deep-learning based models)
which typically hold the network weights in-memory, are instantiated as standalone, stateless microservices. They expose a
lightweight REST API, which is consumed only from the workers.
The migration to the new deployment architecture was driven by several advantanges compared to the
previous behavior, which was calling the model inference in-line on the workers.
First, a particular model microservice can better exploit node-local thread parallelism and memory-sharing
for optimised vertical scaling.
Second, the memory demand for each model instance is well-predictable (a static amount for the model-state and a
buffer for one batch of input pages) and therefore can be accounted for in the Kubernetes cluster resource management.
Third, the infrastructure does not have to provision the memory for serving multiple models in different
chunks, since each model can be scheduled independently.
Fourth, the horizontal auto-scaling components of Kubernetes allow to scale up and down the number of
replicas to ensure fast and stable operation of each model independently.
Fifth, the encapsulation of each type of model into a separate container avoids conflicts between runtime library
dependencies (e.g. running on different versions of torch~\cite{NEURIPS2019_9015} or separating environments of different libraries).
Sixth, removing code-dependencies between models and the logic of the internal pipeline makes it trivial to
customise which models need to run on a particular document collection and new model types can be added
without requiring any code changes in the underling platform.

The CCS platform is configured and deployed into a Kubernetes cluster through Helm, which
encapsulates all system configuration into a chart. It also allows linking to managed persistence services (such as MongoDB\- and COS)
offered by the respective cloud providers.
The presented \textit{seq2seq} model was integrated into CCS as a new
microservice, which is achieved by building a specific container image bundling the ML runtime, inference code and pre-trained
model weights, and then adding respective deployment configurations to CCS to launch it as a microservice.
For development purposes, the container can be built and deployed independently from CCS and then registered at runtime in the CCS Web UI. We developed Continuous Integration pipelines to ensure that building and updating
the model microservices with new code or data is seamless and easy to maintain.

The new model is commonly served on a CPU-only environment with an allocation of 180MB of memory per
replica and a dynamic varying number of replicas ranging from one to ten based on the load. This setup
allows to support a CCS processing throughput of 1 pages per second already with the single replica and
up to 10 pages per second with 10 replicas.

\section{Conclusion}

We have presented a model based on recursive neural networks, which can accurately classify each PDF
printing command according to its structural function in the document. To the best of our knowledge,
this is the first time that a deep neural network has been applied in this way to address the PDF
conversion problem. We have shown that this approach yields more fine-grained, precise output than
current state-of-the-art models. It is also more accurate (in terms of $F_1$ scores) and
computationally more efficient (both in time-to-solution and in memory footprint) than image-based
approaches.

We have described the iterative process of developing this model, including important lessons learned.
The most performant methods in the literature may not always be the most suitable for our problem and our
available data, like we observed with the transformer-based networks. Moreover, data preparation and feature
engineering lead to significant improvements in accuracy and overall complexity of the model.

Finally, we showed the advantages of deploying the model as a microservice in a cloud-native application.
With the integration in an automated pipeline with distributed, scalable workloads, we can maximize the
throughput of available resources and ensure an efficient deployment process. As such, the model and application
presented in this paper qualify very well for converting large and diverse document collections such as arXiv.

\section{Broader Impact}

The recent COVID-19 pandemic has highlighted again that fast, scalable document conversion tools are
necessary to make the latest research insights discoverable. At the time of writing, more than
50,000 PDF documents related to the subject have been published on various online archives
(primarily BioRxiv, MedRxiv and ChemRxiv) as well as provided by various publishers (e.g.~Elsevier
and Wiley). These large document corpora contain many different layouts and must be converted with
the best possible accuracy due to their highly technical content. As such, the model we have
developed to classify individual PDF printing commands is of tremendous value and can be used
directly in applications that make COVID-19-related literature searchable. A prime example of such
an application is the IBM COVID-19 Deep Search
platform\footnote{\url{https://www.research.ibm.com/covid19/deep-search/}}, which was made available
to all researchers and scientists across the world.

The broader impact of this work is twofold. On the one hand, it is absolutely imperative that new AI
models be developed that can structure and extract the data contained in PDF documents. A crucial
step in this process is the accurate conversion of PDF documents into formats that can express the
structure of the document, such as JSON. The new structure recovery model we present in this paper
addresses this problem with great success. On the other hand, this paper demonstrates how recursive
neural networks can be used successfully in rather specialized, non-mainstream application domains
(PDF structure recovery is not nearly as prevalent as image classification, object detection or
machine translation tasks, for example). We intended to present not only the successful models, but
also the failures. The purpose of showing the successes and failures is to guide future research
with direct examples of what may and may not work.

\bibliography{acmart}

\end{document}